\documentclass{article}

\usepackage{microtype}
\usepackage{graphicx}
\usepackage{booktabs} 

\usepackage[hyphens]{url}
\usepackage{hyperref}

\usepackage[accepted]{icml2021}

\usepackage{bm}
\usepackage{amsmath}
\usepackage{amsthm}
\usepackage{algorithm}
\usepackage{algorithmic}
\usepackage{siunitx}
\usepackage{makecell}
\usepackage{multirow}
\usepackage{amssymb}
\usepackage{diagbox}
\usepackage{balance}
\usepackage{xspace}
\usepackage{color}
\usepackage[subrefformat=parens,labelformat=parens,margin=20pt]{subfig}
\graphicspath{{img/}}
\usepackage{comment}

\def\ie{\textit{i.e.}\xspace}

\def\eg{\textit{e.g.}\xspace}
\def\etal{\textit{et~al.}\xspace}

\newcommand{\vect}[1]{\bm{#1}}
\newcommand{\bacc}[1]{\underline{#1}}
\newcommand{\brob}[1]{\textbf{#1}}

\newcommand{\scheme}{CARTL\xspace}

\icmltitlerunning{\scheme: Cooperative Adversarially-Robust Transfer Learning}

\begin{document}

\twocolumn[
\icmltitle{\scheme: Cooperative Adversarially-Robust Transfer Learning}



\icmlsetsymbol{equal}{*}

\begin{icmlauthorlist}
\icmlauthor{Dian Chen}{whu}
\icmlauthor{Hongxin Hu}{buffalo}
\icmlauthor{Qian Wang}{whu}
\icmlauthor{Yinli Li}{whu}
\icmlauthor{Cong Wang}{cityu}
\icmlauthor{Chao Shen}{xjtu}
\icmlauthor{Qi Li}{tsinghua}

\end{icmlauthorlist}

\icmlaffiliation{whu}{School of Cyber Science and Engineering, Wuhan University, Wuhan 430072, Hubei, China}
\icmlaffiliation{buffalo}{Department of Computer Science and Engineering, University at Buffalo, Buffalo, NY 14260, USA}
\icmlaffiliation{cityu}{Department of Computer Science, City University of Hong Kong, HK SAR, China}
\icmlaffiliation{xjtu}{School of Cyber Science and Engineering, Xi'an Jiaotong University, Xi'an 710049, Shanxi, China}
\icmlaffiliation{tsinghua}{Institute for Network Sciences and Cyberspace \& BNRist, Tsinghua University, Beijing 100084, China}

\icmlcorrespondingauthor{Qian Wang}{qianwang@whu.edu.cn}

\icmlkeywords{transfer learning, adversarial examples, adversarial learning}

\vskip 0.3in
]



\printAffiliationsAndNotice{} 

\begin{abstract}
Transfer learning eases the burden of training a well-performed model from scratch, especially when training data is scarce and computation power is limited.
In deep learning, a typical strategy for transfer learning is to freeze the early layers of a pre-trained model and fine-tune the rest of its layers on the target domain. 
Previous work focuses on the accuracy of the transferred model but neglects the transfer of adversarial robustness. 
In this work, we first show that transfer learning improves the accuracy on the target domain but degrades the inherited robustness of the target model. 
To address such a problem, we propose a novel cooperative adversarially-robust transfer learning (CARTL) by pre-training the model via \textit{feature distance minimization} and fine-tuning the pre-trained model with \textit{non-expansive fine-tuning} for target domain tasks. 
Empirical results show that CARTL improves the inherited robustness by about 28\% at most compared with the baseline with the same degree of accuracy. 
Furthermore, we study the relationship between the batch normalization (BN) layers and the robustness in the context of transfer learning, and we reveal that freezing BN layers can further boost the robustness transfer.
\end{abstract}

\section{Introduction} \label{sec::introduction}
The immense progress of deep neural networks (DNNs) leads interactions with machines to a new era. In many fields, DNNs achieve high performance, even better than humans. However, training such a model requires a well-designed network architecture, massive high-quality training data, and extensive computational resources. Obviously, it is impractical for small-scale scenarios due to limited GPUs or insufficient training datasets. 

When further implementing DNNs, we are facing more problems. Numerous research efforts have revealed the brittle robustness of DNNs, which hinders their applications in many security-critical scenarios. Previous work on adversarial examples \cite{szegedy_intriguing_2014-1,papernot_limitations_2016-1,moosavi-dezfooli_deepfool_2016-1,carlini_towards_2017-1,kurakin_adversarial_2017} demonstrated that DNNs can be deceived when given the input with a carefully-crafted perturbation. To solve this problem, \textit{adversarial training} \cite{goodfellow_explaining_2015,madry_towards_2018,kannan_adversarial_2018} has been considered as a promising defensive approach for improving the adversarial robustness of DNNs. The key idea of these approaches is to generate adversarial examples during model training and add them to training datasets. An extra computational burden, however, is introduced to the model training process.

Regarding the above prerequisites of model training, prior work \cite{pan_survey_2010,bengio_deep_2012,yosinski_how_2014} proposed \textit{transfer learning} to obtain high-performance DNN models with significantly reduced efforts. 
It can greatly ease the burden in the (adversarially) training process, especially for those with limited capabilities. Thus, it has been considered as a promising machine learning as a service (MLaaS) technique in the industry \cite{microsoft_transfer_learning,google_transfer_learning}.
The idea of transfer learning is similar to the knowledge transfer in the human world, where the knowledge obtained from the \textit{source domain} is applied to the \textit{target domain} for improving model performance. For DNNs, the ``knowledge'' is included in the weights of models. We call the model trained on the source domain the \textit{source model} and the one for the target domain the \textit{target model} \cite{utrera_adversarially-trained_2020}.

So far, most research efforts have mainly been devoted to improving the accuracy of the target model \cite{kornblith_better_2019,utrera_adversarially-trained_2020,salman_adversarially_2020}, but neglecting its robustness.
The most recent work from Shafahi \etal \yrcite{shafahi_adversarially_2020} discussed how the robustness transfers in transfer learning and pointed out that the target model can inherit the robustness from an adversarially pre-trained source model. However, it suggested only fine-tuning the \textit{last} fully-connected layer for inheriting robustness, which fails to cover more general scenarios where the target model requires to fine-tune \textit{multiple} layers in transfer learning \cite{wang_great_2018,utrera_adversarially-trained_2020}.


In this work, we first provide a complete evaluation for the robustness transfer and demonstrate that the robustness transfer is highly affected by the transfer strategy, namely, the number of fine-tuned layers during transfer learning. Specifically, we transfer a robust source model, which is adversarially trained on the source domain, to the target domain while freezing its first few layers. Our evaluation indicates that as the number of fine-tuned layers increases, the target model's accuracy also improves. However, its robustness only improves at the beginning but soon starts to decrease afterward.

\begin{figure}
    \center
    \includegraphics[width=\linewidth]{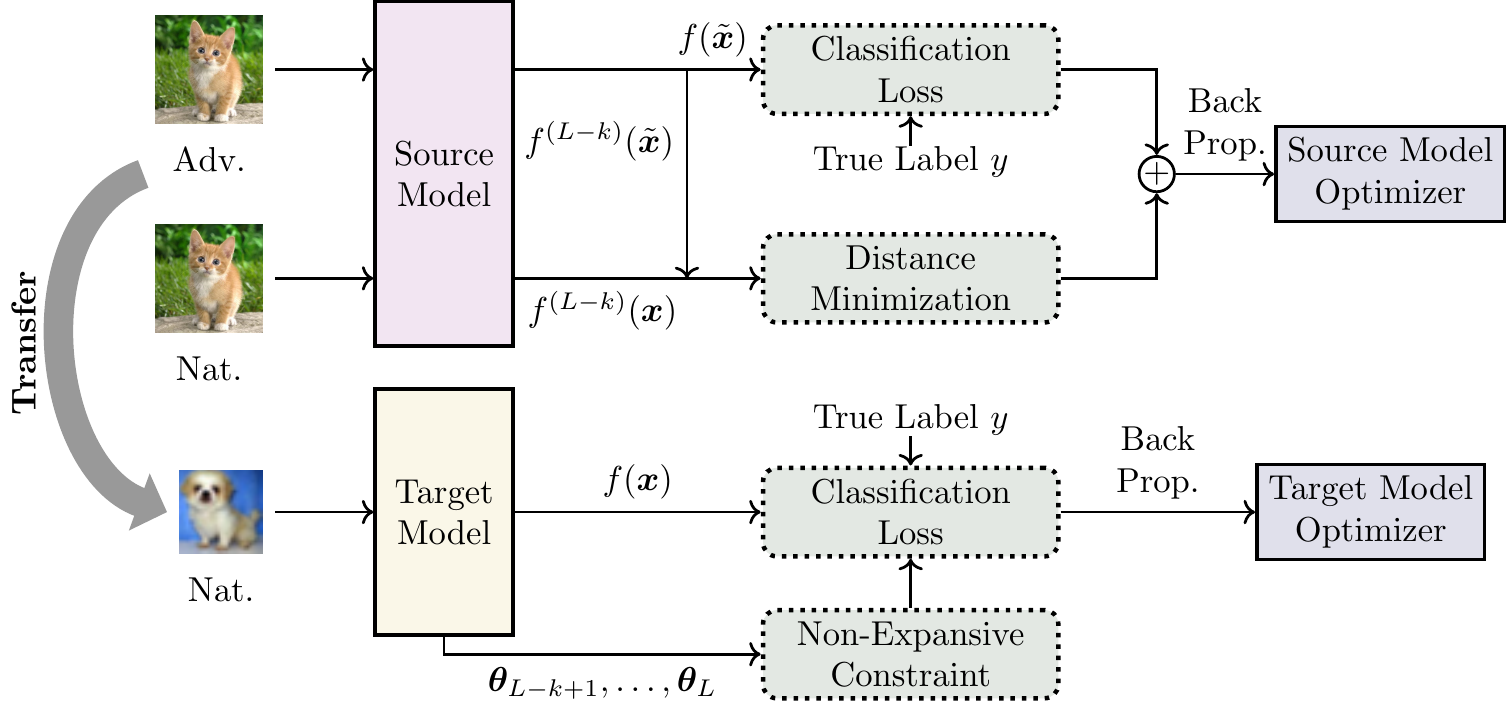}
    \caption{Overview of \scheme.}
    \label{fig::overview}
\end{figure}

Based on the above observation, we find that training more layers enables the target model to adapt to the target domain, which is good for improving accuracy on the target domain. At the same time, modifying more early layers of the source model, which can be seen as a robust feature extractor, reduces the robustness of the target model. To improve this trade-off, we present a new approach, cooperative adversarially-robust transfer learning (\scheme), as illustrated in Figure \ref{fig::overview}. 
In \scheme, we consider training a robust feature extractor of the source model, which outputs similar features when given natural inputs and corresponding adversarial inputs, and we call this \textit{feature distance minimization}. In addition, to further reduce the negative effects caused by the feature differences between natural and adversarial inputs, we propose \textit{non-expansive fine-tuning}, which is used to control the Lipschitz constant of the network during fine-tuning. We call our method \textit{cooperative} because it consists of an adjusted adversarial training on the source model side and a constrained fine-tuning on the target model side.
We emphasize that \scheme requires \textit{no adversarial examples} during fine-tuning, which is far more efficient than adversarial training.

Besides, we conduct a study, for the first time to our best knowledge, on the relationship between the batch normalization (BN) layers and the robustness in the context of transfer learning. Our results reveal that selectively freezing BN layers' parameters helps boost the robustness transfer. 
We conduct extensive experiments on several transfer learning scenarios and observe that the target model freezing affine parameters of BN layers obtains higher robustness with negligible loss of accuracy. We also show that though BN layers' statistics play  a crucial role in the robustness transfer, it will degrade the target model's accuracy.

We summarize our main contributions as follows:
\begin{itemize}
    \item Through experimental analysis, we reveal that there is a trade-off between accuracy and robustness during transfer learning, which has been overlooked by prior work. Specifically, the target model obtains higher accuracy on the target domain as it fine-tunes more layers. However, as the number of fine-tuned layers increases, the target model’s robustness is greatly affected and eventually severely degraded.
    \item We propose a new transfer learning strategy, \scheme, for improving the accuracy-robustness trade-off of the target model. Our experimental evaluations on broadly-used datasets show that our design improves the target model's inherited robustness while gaining competitive accuracy on the target domain.
    \item We also conduct extensive experiments on several transfer learning scenarios to demonstrate that selectively freezing the BN layers can further boost the robustness transfer. 
\end{itemize}

\section{Related Work} \label{sec::related_works}
Various techniques focusing on defending adversarial examples have been proposed \cite{papernot_distillation_2016-1,papernot_extending_2017,xie_feature_2019,song_pixeldefend_2018}. However, many defenses were still proven to be vulnerable to stronger attacks \cite{carlini_towards_2017-1,athalye_obfuscated_2018-1,tramer_adaptive_2020}. Despite the fails of many defenses, adversarial training \cite{szegedy_intriguing_2014-1,kurakin_adversarial_2017,madry_towards_2018,kannan_adversarial_2018} is still widely regarded as a promising defense for protecting trained models from adversarial examples and has been extensively discussed \cite{tramer_ensemble_2018,schmidt_adversarially_2018,tsipras_robustness_2019,tramer_adversarial_2019,zhang_theoretically_2019}.

Goodfellow \etal \cite{goodfellow_explaining_2015} firstly observed that adding adversarial examples into the training datasets improves adversarial robustness, and the strategy is called adversarial training. Madry \etal \cite{madry_towards_2018} used a strong attack method to generate adversarial examples during training, which demonstrated that the trained model is robust to single-step attacks as well as multi-step attacks. 
So far, compared with the standard training, the main drawback of adversarial training is the degraded efficiency of training, which is introduced by the generation of adversarial examples. Meanwhile, some literature \cite{shafahi_adversarial_2019,zhang_you_2019,wong_fast_2020} works on efficiency optimization for adversarial training.
Very recently, Shafahi \etal \cite{shafahi_adversarially_2020}  gave an evaluation of adversarial robustness in transfer learning. They found that the target model can efficiently inherit the adversarial robustness from an adversarially pre-trained model. However, they only considered the target model that fine-tunes the \textit{last} layer of the source model while ignoring fine-tuning more layers of the source model.

Similar to adversarial training, other work \cite{cisse_parseval_2017,qian_l2-nonexpansive_2019,lin_defensive_2019-1} improves the robustness of the model during the model-training stage with a different idea. The seminal work of Szegedy \etal~\cite{szegedy_intriguing_2014-1} attributed the vulnerability of adversarial examples to the instability of the model, which can be mitigated by controlling the Lipschitz constant of the model.
Based on this idea, Cisse \etal \cite{cisse_parseval_2017} proposed to add a regularization term during model training to constrain the Lipschitz constant of the entire model. The Lipschitz constant of the entire model approximates to one, which makes the model's final prediction less sensitive to little perturbations.
The following work \cite{qian_l2-nonexpansive_2019} relaxed the training limitations of Cisse \etal \cite{cisse_parseval_2017}, which forces the weight matrix of each layer to be orthogonal, providing more freedom for training.


We emphasize that our work is orthogonal to prior work on domain adaptation \cite{shu_dirt-t_2018}. In this work, we focus on transferring the adversarial robustness from the source domain to the target domain, while domain adaptation refers to leveraging the source model's knowledge to improve the accuracy of the target model. Besides, recent work~\cite{utrera_adversarially-trained_2020,salman_adversarially_2020} also reveals that an adversarially pre-trained model tends to improve the target model's accuracy. We leave the studies of the connection between the source domain robustness and the target domain accuracy for our future work.

\section{Preliminary} \label{sec::preliminary}
We consider DNN-based classification tasks and define an $L$-layer feed-forward DNN model:
\begin{equation}
    f(\cdot; \vect{\theta}) :=  \left(f^{L}_{\vect{\theta}_L} \circ f^{L-1}_{\vect{\theta}_{L-1}} \circ \cdots f^{1}_{\vect{\theta}_1} \right)(\cdot),
    \label{eq::layer_wise_model}
\end{equation}
which is parameterized by $\vect{\theta}:=\{\vect{\theta}_1, \ldots, \vect{\theta}_{L}\}$. We use $f^{k}$ to represent the $k$th layer of the model $f$ and use $f^{(k_1..k_2)}$ to represent layers ranging from $k_1$ to $k_2$, \ie, $f^{(k_1..k_2)} := f^{k_2} \circ \cdots \circ f^{k_1}$. We also denote the first $k$ layers as $f^{(k)}$ for shorthand.

To train the model, given a proper loss function $\mathcal{L}$, \eg, cross-entropy loss, we want to find the optimal parameters $\vect{\theta}^*$ that minimizes the risk:
\begin{equation}
    \arg \min_{\vect{\theta}} \mathbb{E}_{(\vect{x}, y) \sim \mathcal{D}} \left[\mathcal{L} (f(\vect{x}; \vect{\theta}), y) \right],
\end{equation}
where $\mathcal{D}$ is the data distribution of image-label pair $(\vect{x}, y)$. We define data $\vect{x} \in [0,1]^d$, where $d$ is the input dimension, and define label $y \in \{0, 1,\ldots, C-1\}$, where $C$ is the number of class labels. 

\subsection{Adversarial Examples \& Adversarial Training} \label{sec::preliminary::ae_and_at}
Most adversarial example attacks consider an $\ell_p$-norm constrained optimization problem that can be generalized as:
\begin{equation}
    \begin{gathered}
        \arg \max_{\vect{\delta}} \  \mathcal{L}\left( f(\vect{x}+\vect{\delta}; \vect{\theta}), y \right) \quad s.t. \  \parallel \vect{\delta} \parallel_p \leq \epsilon.    
    \end{gathered}
    \label{eq::adversarial_examples}
\end{equation}
The hyper-parameter $\epsilon$ guarantees that the perturbation $\vect{\delta}$ is imperceptible. In our work, we consider $\ell_{\infty}$-norm-based attacks \cite{madry_towards_2018} and let $\epsilon = 8 / 255$.

We follow \cite{madry_towards_2018} defining the adversarial training as a saddle-point problem that aims to minimize a variant of the training risk:
\begin{equation}
        \arg \min_{\vect{\theta}} \mathbb{E}_{(\vect{x}, y) \sim \mathcal{D}} \left[
        \max_{\parallel \vect{\delta} \parallel_p \leq \epsilon} \  \mathcal{L} \left(f \left(\vect{x}+\vect{\delta}; \vect{\theta} \right), y \right)
    \right].
    \label{eq::adversarial_training}
\end{equation}
The inner maximization problem can be approximated by an iterative version of Eq. \eqref{eq::adversarial_examples}, known as the projected gradient descent (PGD). The method can be summarized as:
\begin{equation}
    \begin{gathered}
        \vect{\delta}^{i+1} := 
        \Pi\left(
            \vect{\delta}^{i} + \alpha \cdot \text{sign}\left(
                \nabla_{\vect{\delta}} \mathcal{L}\left(f\left(\vect{x} + \vect{\delta^{i}}\right), y\right)
            \right)
        \right),
    \end{gathered}
    \label{eq::pgd}
\end{equation}
where $\alpha$ is the step size. For $\ell_\infty$-based perturbations, the projection $\Pi$ clips the noise $\vect{\delta}$ to the interval $[-\epsilon, \epsilon]$.
Compared with single-step attacks, PGD achieves higher error rates since it tends to find the global maxima of Eq.  \eqref{eq::adversarial_examples}. In the following sections, we represent the $N$-step PGD attack as PGD-$N$.

\subsection{Lipschitz Constant} \label{sec::preliminary::lip_const}
The Lipschitz constant defines an upper bound of the function's slope. If we can find such an upper bound, we call the function Lipschitz continuous, and formally, it can be described as
\begin{equation}
    \parallel f(x) - f(x') \parallel_2 \leq \Lambda \cdot \parallel x - x' \parallel_2, 
    \label{eq::lipschitz_constant}
\end{equation}
where $\Lambda$ is the Lipschitz constant. Specifically, we call a function \textit{non-expansive} when $\Lambda \leq 1$. Intuitively, if a function is non-expansive, the deviation of its output is no more than the perturbation on its input.
Recall Eq. \eqref{eq::layer_wise_model}, a DNN model is stacked in a layer-wise manner. From the Lipschitz continuity perspective, we have
\begin{equation}
    \parallel f(\vect{x}) - f(\vect{x}') \parallel_2 \leq \Lambda_L \cdot \Lambda_{L-1} \cdots \Lambda_1 \parallel \vect{x} - \vect{x}' \parallel_2,
    \label{eq::lipschitz_model}
\end{equation}
where $\Lambda_i$ is the Lipschitz constant of the $i$th layer.
We can observe that the Lipschitz constant of the whole model is a product of each layer's Lipschitz constant. When the layers' Lipschitz constants are more than one, a little change of the input may be amplified during forward propagation and result in misclassification \cite{lin_defensive_2019-1}.
The above inequality also implies that we can mitigate the vulnerability of adversarial examples by constraining the Lipschitz constant of each layer no more than one \cite{cisse_parseval_2017,qian_l2-nonexpansive_2019,lin_defensive_2019-1}.

\begin{figure}[t]
    \center
    \includegraphics[width=0.9\linewidth]{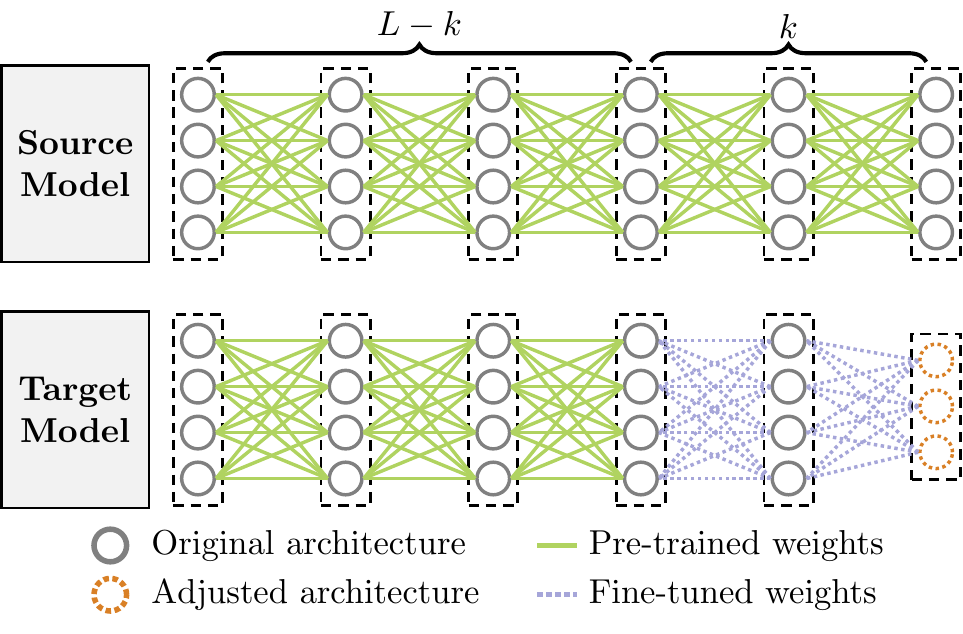}
    \caption{Illustration of transfer learning. 
    }
    \label{fig::transfer_learning}
\end{figure}

\subsection{Transfer Learning} \label{sec::preliminary::tl}
The main idea of transfer learning is to transfer ``knowledge'' 
from a pre-trained source model to a target model for solving target domain tasks. In deep learning, a widely-adopted method for transfer learning is that the target domain tasks copies the whole \textit{pre-trained} model from the source model, refines the architecture, typically adjusting the last fully-connected layer, and fine-tunes the last $k \in \{1, \ldots, L\}$ layers \cite{wang_great_2018,utrera_adversarially-trained_2020}. Formally, it can be formulated as
\begin{equation}
    \arg \min_{\bar{\vect{\theta}}} \mathbb{E}_{(x,y)\sim\mathcal{D}} \left[
        \mathcal{L}\left(
            f_{\bar{\vect{\theta}}}^{(L-k + 1 .. L)}
            \left(f^{(L-k)}(\vect{x})\right), y 
        \right)
    \right],
\end{equation}
where $\bar{\vect{\theta}}:=\{\vect{\theta}_{L-k+1}, \ldots, \vect{\theta}_{L}\}$. An illustration of transfer learning is depicted in Figure \ref{fig::transfer_learning}. Intuitively, we can view the output of the first frozen $L-k$ layers as the extracted features of the input $\vect{x}$, while the fine-tuned part of the target model is a sub-model that directly takes as input those features. During transfer learning, we reuse the powerful feature extractor of the source model (green solid lines) and adapt the sub-model (blue dashed lines) for the target domain task.


\begin{figure*}[ht]
    \centering
    \subfloat{ \includegraphics[width=0.46\linewidth]{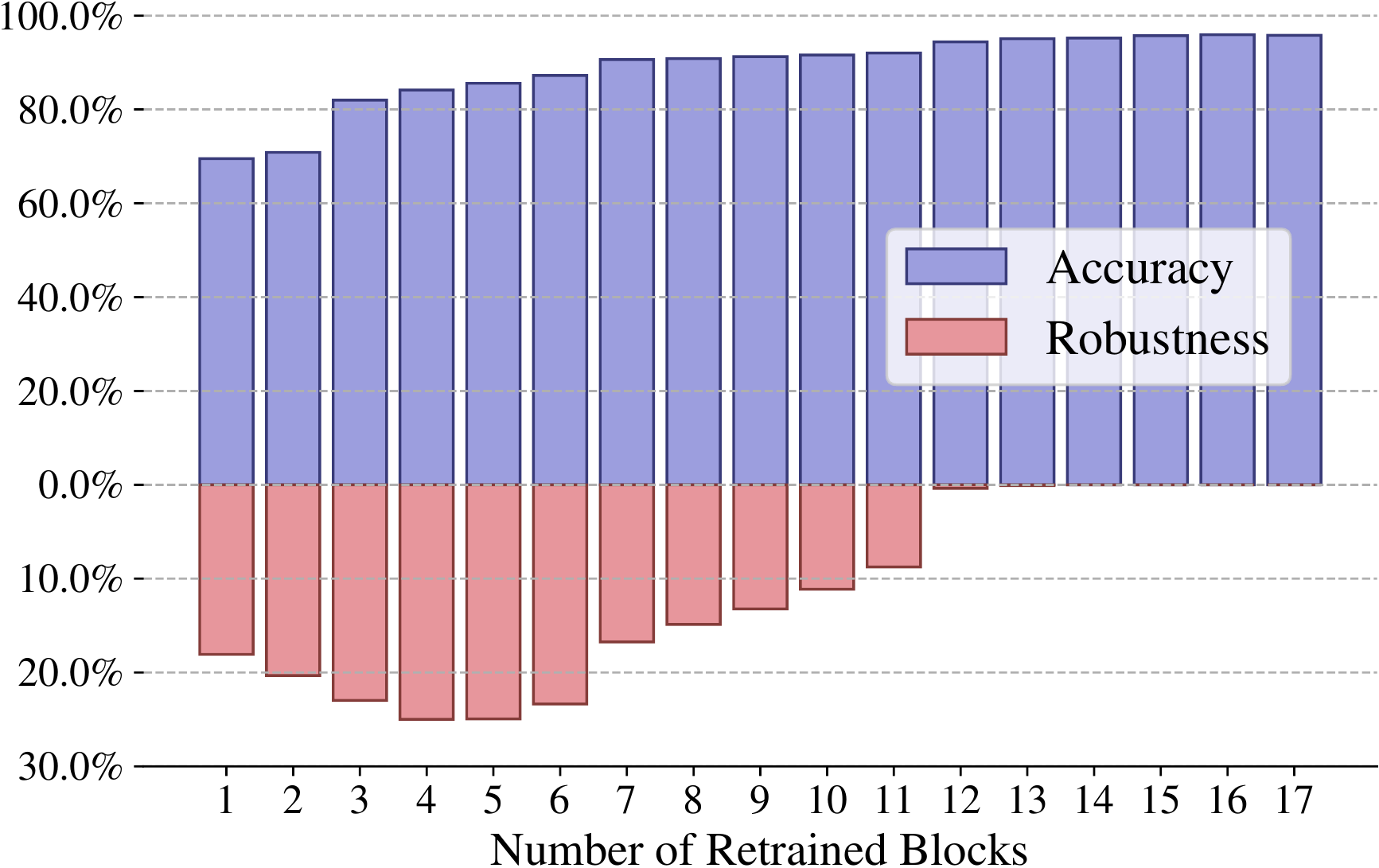} }
    \subfloat{ \includegraphics[width=0.46\linewidth]{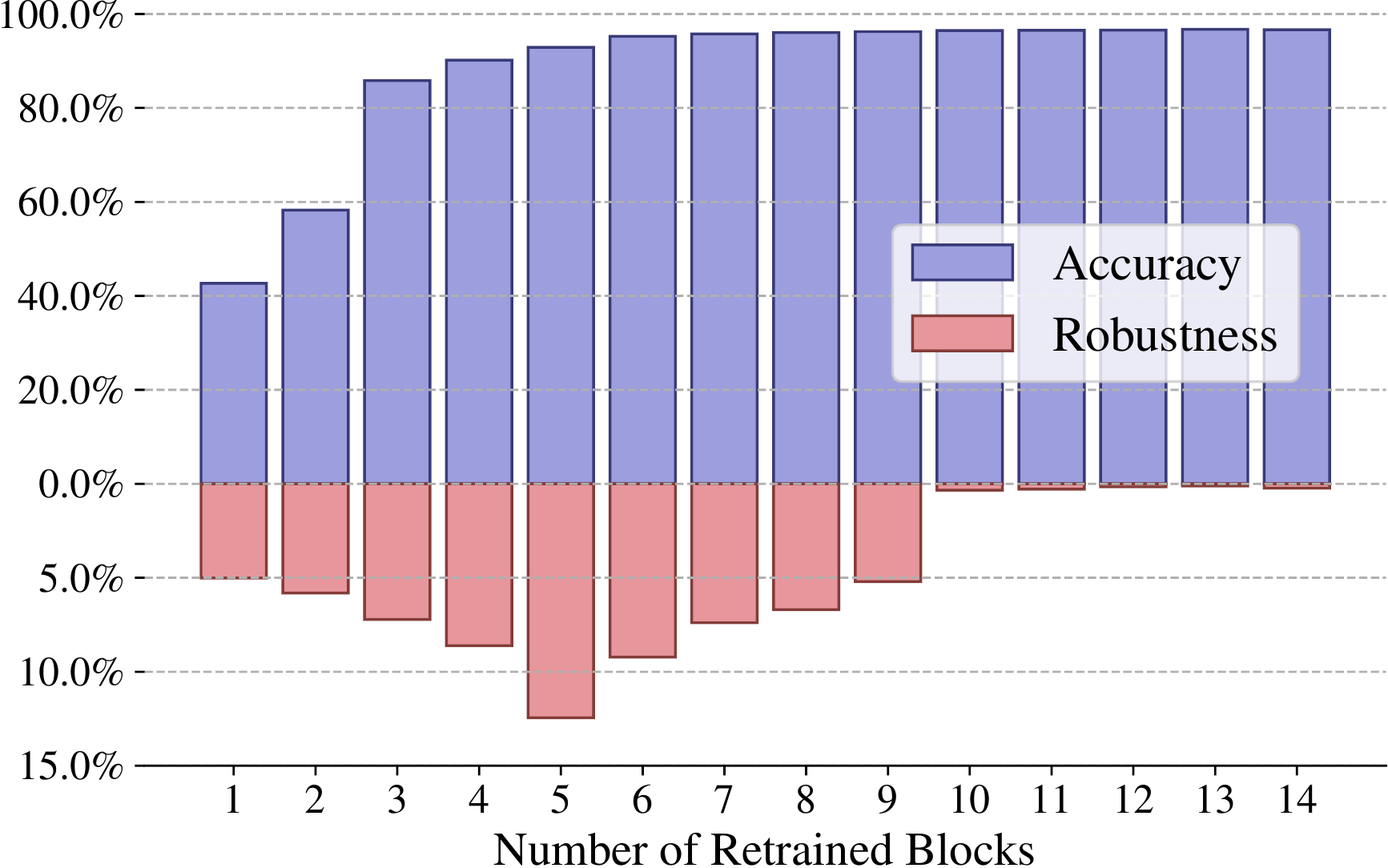} }
    \caption{ Accuracy and robustness of target models transferred from robust source models. \textit{Left}: A WRN 34-10 model is transferred from CIFAR-100 to CIFAR-10. \textit{Right}: A WRN 28-4 model is transferred from CIFAR-10 to SVHN. }
    \label{fig::problem_define}
\end{figure*}

\section{Problem Statement} \label{sec::prob_state}
The previous work of Shafahi \etal \yrcite{shafahi_adversarially_2020} gave an empirical analysis of robustness transfer. However, their exploratory experiments put less attention on the accuracy. Besides, they mainly considered a strategy that all layers but the \textit{last} fully-connected layer are frozen, which does not cover more general cases. In this work, we study how both the robustness and accuracy transfer while the last $k$ layers are fine-tuned. 

To see the effect of fine-tuning on the robustness and accuracy, we adversarially train a Wide-ResNet (WRN) 34-10 \cite{zagoruyko_wide_2017} on CIFAR-100 and a WRN 28-4 on CIFAR-10 as source models, then transfer them to CIFAR-10 and SVHN, respectively.
The source models are trained with PGD-$7$, and the perturbation is constrained in an $\ell_\infty$ ball with a radius of $\epsilon=8/255$. During transferring, we break the source models into blocks and fine-tune them in the unit of blocks (\eg, two layers at once for a WRN block). Then we report the adversarial robustness of the target models against the PGD-100 attack. We emphasize that our settings are different from the exploration in \cite{shafahi_adversarially_2020}, where the last $k$ blocks were instead fine-tuned on the source domain.

Figure \ref{fig::problem_define} illustrates how both the accuracy and robustness are affected during transfer learning. As can be seen, only retraining the last fully-connected layer fails to guarantee high accuracy for target domain tasks. Besides, the insufficient accuracy also results in lower robustness. If we further fine-tune the last few layers, the model accuracy is increased together with the increased robustness. We attribute this phenomenon to the increment of the accuracy of natural inputs. The accuracy is continuously increased while we fine-tune more layers, but the robustness quickly drops and ends with negligible. The results demonstrate there is a trade-off between the target model's accuracy on the target domain and its robustness inherited from the source model. Besides, simply fine-tuning the last few layers does help the target model inherit the accuracy and robustness in a low cost.

Furthermore, besides the above observations, it is natural to raise another question: 
\begin{center}
    \textit{ Can the target model obtain high accuracy while inheriting more robustness from the source model? } 
\end{center}
To answer with this question, we propose a novel strategy for transfer learning, which further improves the accuracy-robustness trade-off during transfer learning.

\section{Our Design: \scheme} \label{sec::design}
In this section, to cope with the question raised in Section \ref{sec::prob_state}, we propose a new approach, cooperative adversarially-robust transfer learning (\scheme), to improve both the robustness and accuracy of the target model. We divide layers of a robust source model into two parts during transfer learning according to whether they will be retrained. 
We take the frozen part as a feature extractor like~\cite{utrera_adversarially-trained_2020}, and we propose that it outputs similar features if given natural examples and corresponding adversarial examples. As for the trainable part fine-tuned on the target domain, we aim to reduce the classification error caused by the differences between the features extracted from natural and adversarial inputs. Without loss of generality, we assume that the target model fine-tunes the last $k$ layers and freezes the first $L-k$ layers during transfer learning. We introduce our scheme, starting with training the source model.

\subsection{Feature Distance Minimization} \label{sec::design::fdm}
Our intuition is that if a specific layer of a model extracts similar features from two inputs, the subsequent layers tend to classify the inputs into identical classes, even they have different labels \cite{wang_great_2018}.
Hence, we consider making the frozen part of the source model a robust feature extractor that can output similar features given natural examples and the corresponding adversarial examples. 
To do so, we propose \textit{feature distance minimization} (FDM) to reduce the dissimilarity of the extracted features.
Specifically, for the first $L-k$ layers that take as input $\vect{x}$ and output intermediate features $f^{(L-k)}(\vect{x})$, FDM adds a penalty term to the training loss for two different inputs $\vect{x}$ and $\tilde{\vect{x}}$:
\begin{equation}
    \begin{aligned}
        \mathcal{L}_{AT} + \lambda \cdot D\left(f^{(L-k)}(\vect{x}), f^{(L-k)}(\tilde{\vect{x}})\right).
    \end{aligned}
    \label{eq::fdm_penalty}
\end{equation}
Here, $\lambda$ is the hyper-parameter controlling the strength of the FDM penalty term, and $D$ is a distance metric measuring the dissimilarity between two features. $\mathcal{L}_{AT}$ is the loss function used during adversarial training, and $\tilde{\vect{x}}$ is the adversarial example corresponding to the natural example $\vect{x}$.

Intuitively, we can view extracted features as points in a subspace of $\mathbb{R}^d$, where $d$ is the feature's dimension. For a source model that can extract similar intermediate features, features of natural examples and adversarial examples should be close enough. Thus, we propose to use the Euclidean distance between the natural feature and adversarial feature as the penalty term. Specifically, we adjust the original training loss Eq. \eqref{eq::adversarial_training} to
\begin{equation}
    \begin{aligned}
        \mathcal{L}_{AT} + \frac{\lambda}{\sqrt{d}} \cdot \sum \parallel f^{(L-k)}(\vect{x}) - f^{(L-k)}(\tilde{\vect{x}})\parallel_2.
    \end{aligned}
    \label{eq::l2_fdm}
\end{equation}

\begin{table*}[ht!]
    \center
    \caption{ Effect of selectively freezing BN layers in various scenarios. The third and fourth columns are results of target models freezing affine parameters of the feature extractor. The fifth and sixth rows are results of target models freezing \textit{all} parameters of the feature extractor. For each transfer learning scenario, the first rows are results of target models fine-tuning \textit{all} parameters of the sub-model, and the second rows are those that freeze affine parameters. }
    \begin{tabular} {c c c c c c}
        \toprule
        \multirow{2}{*}{} & \multirow{2}{*}{} & \multicolumn{2}{c}{$\vect{W}, \vect{b}$} & \multicolumn{2}{c}{$\vect{\mu}, \vect{\sigma}, \vect{W}, \vect{b}$} \\
        
                                                                      \cmidrule(lr){3-4}              \cmidrule(lr){5-6} 
        &                                                           & Acc.(\%)      & Rob.(\%)      & Acc.(\%)      & Rob.(\%)  \\
        \midrule
        \multirow{2}{*}{ CIFAR-100 $\rightarrow$ CIFAR-10 ($k=8$) }
        &   -                                                       & \bacc{91.17}  & 14.36         & 90.86         & 14.89         \\
        &   $\vect{W}, \vect{b}$                                    & 90.70         & 17.41         & 90.84         & \brob{18.54}  \\

        \midrule
        \multirow{2}{*}{CIFAR-10 $\rightarrow$ GTSRB ($k=6$) }
        &   -                                                       & \bacc{93.02}  & 30.22         & 89.29         & 32.22         \\
        &   $\vect{W}, \vect{b}$                                    & 92.13         & 32.22         & 88.94         & \brob{34.53}  \\

        \midrule
        \multirow{2}{*}{CIFAR-10 $\rightarrow$ SVHN ($k=6$) }
        &   -                                                       & \bacc{95.29}  &  3.88         & 95.24         &  9.22         \\
        &   $\vect{W}, \vect{b}$                                    & 95.16         &  4.90         & 94.86         & \brob{11.52}  \\

        \midrule
        \multirow{2}{*}{CIFAR-10 $\rightarrow$ SVHN ($k=5$) }
        &   -                                                       & \bacc{93.47}  &  4.71         & 92.92         & 12.45         \\
        &   $\vect{W}, \vect{b}$                                    & 93.41         &  5.64         & 92.10         & \brob{14.16}  \\
        
        \bottomrule
    \end{tabular}
    \label{tab::bn_effect}
\end{table*}

\subsection{Non-Expansive Fine-tuning} \label{sec::design::neft}
The source model trained with FDM outputs similar features if both natural and adversarial examples are given. However, considering the previous discussion in Eq. \eqref{eq::lipschitz_model}, the little dissimilarity between the features may be still amplified during propagation in the rest of the network, leading to misclassification. To improve the target model's robustness, we propose to suppress such the amplification effect via controlling the network's Lipschitz constant.



We start with the basic linear layer that can be expressed as $\vect{x}^{l+1} = \vect{W}^{l}\vect{x}^{l} + \vect{b}^{l}$, where $\vect{W}^{l} \in \mathbb{R}^{d_{out} \times d_{in}}$ and $\vect{b}^{l} \in \mathbb{R}^{d_{out}}$. It is straightforward that the bias $\vect{b}^{l}$ does not affect the Lipschitz constant. Thus, the Lipschitz constant of the linear layer is determined by $\vect{W}^{l}$. Since the Lipschitz constant is upper bounded by the spectrum norm of $\vect{W}^{l}$, \ie, its maximum singular value \cite{szegedy_intriguing_2014-1}, similar to the spectrum normalization \cite{miyato_spectral_2018}, we divide weights $\vect{W}^{l}$ of last $k$ layers by their spectrum norm $\sigma(\vect{W}^{l})$ as
\begin{equation}
    \vect{W}^{l}_{*} := \beta \cdot \frac{\vect{W}^{l}}{\sigma(\vect{W}^{l})}, \quad (l = {L - k + 1, \ldots, L}).
    \label{eq::neft_penalty}
\end{equation}
We also emphasize that different from the naive spectrum normalization, we add a hyper-parameter $\beta \in (0, 1]$ for further scaling the Lipschitz constant of the fine-tuned part. The idea comes from the observation that parameters trained with FDM tend to have a smaller ($<1$) Lipschitz constant. For the details of $\sigma(\cdot)$, please refer to Appendix \ref{app::sec::spectrum_norm}.

For the convolutional layer, we flatten each filter into a vector with $C_{in} \cdot K \cdot K$ dimensions, where $C_{in}$ is the number of input channels, and $K$ is the size of the convolution kernel. We further stack the vectors forming a $C_{out}$-row matrix, where $C_{out}$ is the number of output channels. Hence, we transform the weight matrix as $\vect{W} \in \mathbb{R}^{C_{out} \times (K^2 \cdot C_{in})}$. 
As for the aggregation layer in the residual network \cite{he_deep_2016}, which adds the predecessor layer's output with the shortcut connection's output, we instead modify it to a convex combination of its inputs \cite{cisse_parseval_2017} and manually set the weights be $1/n$, where $n$ is the number of its inputs.


\section{Rethinking Fine-tuning BN Layers} \label{sec::bn}
During our evaluation, we find a strong connection between the BN layer and the transferred robustness. Before presenting experimental evaluations of \scheme, we first investigate how the BN layer affects the target models' robustness in this section. Specifically, we find that selectively freezing the BN layers of source models improves the transferred robustness of target models  and generally brings little negative impact on their accuracy.

We first simply recap the basis of the BN layer.
In current implementations of the BN layer\footnote{\textit{E.g.}, PyTorch, TensorFlow.}, it usually consists of four parameters, including two running statistics $\vect{\mu}$ and $\vect{\sigma}$, and two affine parameters $\vect{W}$ and $\vect{b}$. Typically, a BN layer can be expressed as 
\begin{equation}
    \begin{gathered}
        BN(\vect{x}) := \vect{W} \cdot \frac{\vect{x} - mean(\vect{x})}{\sqrt{var(\vect{x})+ \varepsilon}} + \vect{b}.
    \end{gathered}
    \label{eq::bn_layer}
\end{equation}
During training, both the running statistics $\vect{\mu}$ and $\vect{\sigma}$ are updated with a momentum based on batch's statistics (\ie, $mean$ and $var$), while $\vect{W}$ and $\vect{b}$ are updated via the gradient descent. During inference, BN layers normalize the activation with running statistics instead of batch's statistics.

To see the effect of BN layers in transfer learning, we divide all BN layers of a source model into two sets according to whether they are in the frozen feature extractor or the fine-tunable sub-model (see Section \ref{sec::preliminary::tl}). In total, we consider four cases regarding the BN layers, including updating or freezing source model's running statistics (\ie, $\vect{\mu}$ and $\vect{\sigma}$) in the frozen feature extractor, and fine-tuning or freezing source model's affine weights (\ie, $\vect{W}$ and $\vect{b}$) in the sub-model. For other cases, we note that both $\vect{W}$ and $\vect{b}$ in the feature extractor are naturally frozen in transfer learning, and we also find that freezing running statistics $\vect{\mu}$ and $\vect{\sigma}$ in the fine-tunable layers makes the target model hard to converge. 

\begin{table*}[t]
    \center
    \caption{Accuracy and robustness of target models transferred to CIFAR-10 under different choices of hyper-parameters, including $\lambda$ for FDM, $\beta$ for NEFT and the number of fine-tuned layers. }
    \begin{tabular} {c l | c c c c c c}
        \toprule
        \multicolumn{2}{c}{}                    & \multicolumn{2}{c}{NEFT $\beta = 1.0$}    & \multicolumn{2}{c}{NEFT $\beta = 0.6$}    & \multicolumn{2}{c}{NEFT $\beta = 0.4$}    \\
                                                \cmidrule(lr){3-4}              \cmidrule(lr){5-6}              \cmidrule(lr){7-8}
        \multicolumn{2}{c}{}                    & Acc.($\%$)    & Rob.($\%$)    & Acc.($\%$)    & Rob.($\%$)    & Acc.($\%$)    & Rob.($\%$)    \\
        \midrule
        \multirow{2}{*}{Case-4}
            & $\lambda=0.01$                    & 86.09         & 25.73         & 86.08         & 27.17         & 85.64         & 28.40         \\ 
            & $\lambda=0.005$                   & 85.41         & 25.75         & 85.47         & 27.14         & 85.51         & 28.47         \\ 
        
        \midrule
        \multirow{2}{*}{Case-6}
            & $\lambda=0.01$                    & 87.78         & 25.58         & 87.92         & 27.27         & 87.96         & 29.60         \\ 
            & $\lambda=0.005$                   & 87.66         & 25.97         & 88.07         & 27.64         & 87.79         & 30.94         \\ 
        
        \midrule
        \multirow{2}{*}{Case-8}
            & $\lambda=0.01$                    & 91.85         & 16.36         & 91.63         & 19.22         & 91.55         & 27.47         \\ 
            & $\lambda=0.005$                   & 91.71         & 17.62         & 91.10         & 21.60         & 91.30         & 29.34         \\ 

        \bottomrule
    \end{tabular}
    
    \label{tab::hp_choice}
\end{table*}

We conduct experiments on several transfer learning scenarios where robust models are \textit{naively} transferred to target domains, and we present the results in Table \ref{tab::bn_effect} (more setups are presented in Appendix \ref{app::sec::settings}). It is shown that though it slightly degrades accuracy, freezing all affine parameters (\ie, $\vect{W}$ and $\vect{b}$) of BN layers can further improve the transferred robustness. For example, the target model's robustness is increased from $14.36\%$ to $17.41\%$, while the accuracy is decreased by $0.47\%$ if we transfer a robust source model from CIFAR-100 to CIFAR-10. Unlike the analysis in Section \ref{sec::prob_state}, where we show that reducing the number of fine-tunable layers harms the accuracy of the target model, we can see that merely freezing the BN layers' affine parameters does not aggressively decrease the accuracy while improving the robustness. In addition, reusing the running statistics of the BN layers in the frozen feature extractor plays a crucial role in robustness transfer, \eg, transferring from CIFAR-10 to SVHN. However, it tends to bring more negative effects to the accuracy, especially when transferring from CIFAR-10 to GTSRB. We note that our findings corroborate the recent studies \cite{xie_intriguing_2019,xie_adversarial_2020}, which argue that the BN layers highly relate to robustness.
        
        


\section{\scheme Evaluation}
In this section, we present the experimental results of \scheme compared with the vanilla method and Shafahi's work \yrcite{shafahi_adversarially_2020}. We conduct a detailed experimental analysis for the scenario of transferring from CIFAR-100 to CIFAR-10. Besides, more scenarios are also tested to demonstrate the generality of \scheme.
We report the robustness of target models under the PGD-$100$ attack. 
For more details about experiment settings, please refer to Appendix \ref{app::sec::settings}, and our codes are available on GitHub\footnote{https://github.com/NISP-official/CARTL}.

\begin{figure}
    \center
    \includegraphics[width=\linewidth]{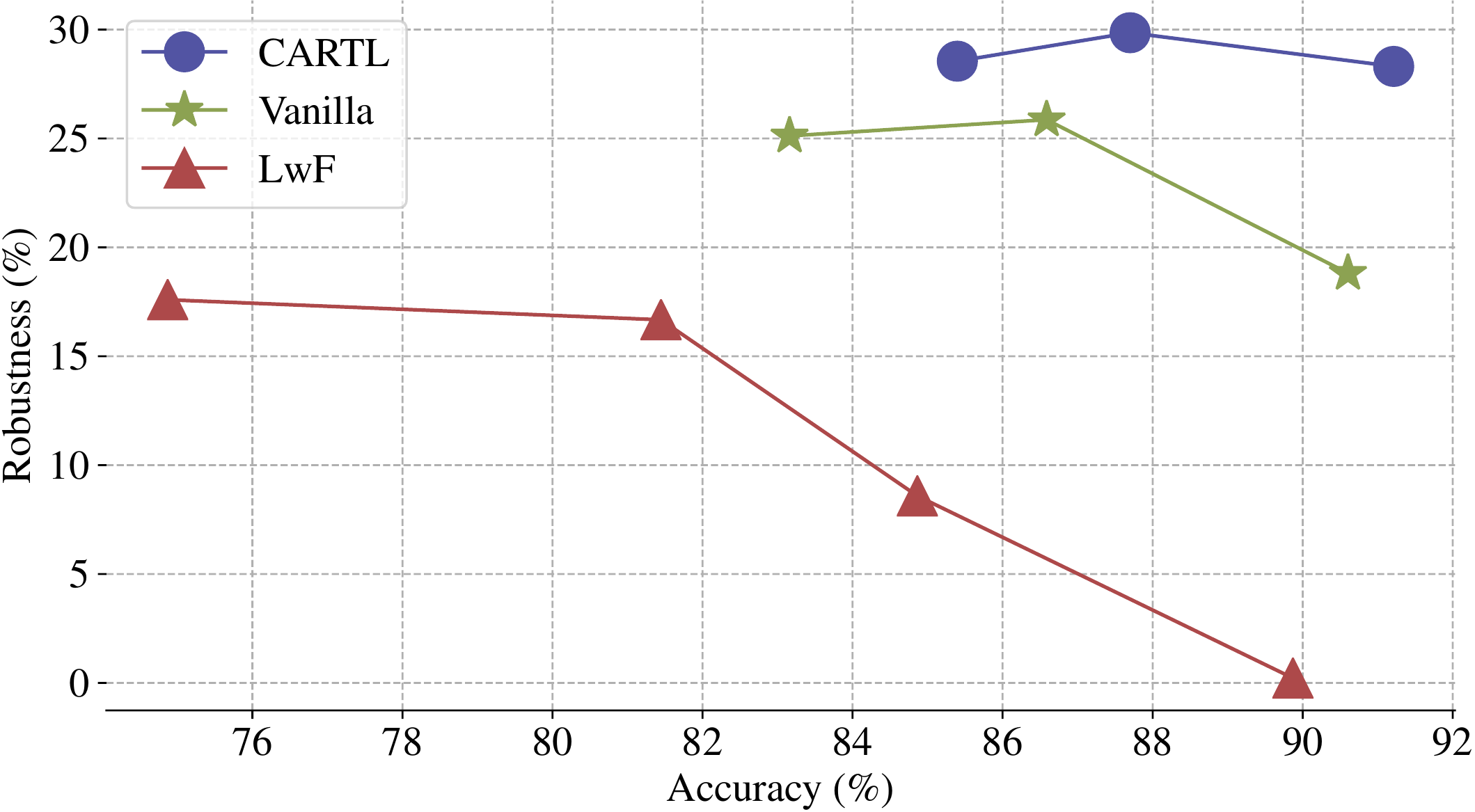}
    \caption{ The robustness-accuracy trade-off of target models transferred with \scheme in comparisons with LwF and the vanilla method when robust models transfer from CIFAR-100 to CIFAR-10. }
    \label{fig::lwf_cmp_ours}
    \vspace{-0.5cm}
\end{figure}

\subsection{Improved Robustness-Accuracy Trade-off}
Shafahi \etal \yrcite{shafahi_adversarially_2020} noticed that merely retraining the last layer maintains the robustness of the source model but results in low accuracy on the target domain. To cope with this problem, they proposed an end-to-end fine-tuning method with learning without forgetting (LwF)\footnote{We give detailed introduction of LwF in Appendix \ref{app::sec::lwf}.}. In addition to LwF, we also consider the vanilla method, which refers to simply fine-tune the last few layers of a robust source model on the target domain without additional techniques. Source models for both LwF and the vanilla method are adversarially trained with \cite{madry_towards_2018}.

Figure \ref{fig::lwf_cmp_ours} qualitatively illustrates the robustness-accuracy trade-off achieved by \scheme in comparison with LwF and the vanilla method. It is shown that to obtain high accuracy, LwF significantly degrades the target model's robustness (close to $0\%$). Moreover, LwF improves the robustness of the target model but aggressively harms its accuracy on the target domain. On the other hand, both the vanilla method and \scheme maintain higher robustness in the case of an equivalent level of accuracy, demonstrating a better robustness-accuracy trade-off. 
Moreover, our method \scheme further improves the accuracy-robustness trade-off on the target domain during transfer learning. Specifically, it improves the robustness by about 28\% compared with LwF when the accuracy is about 90\%.

We also notice that a peak point appears in both curves of CARTL and Vanilla. This phenomenon is in line with the observations in Section \ref{sec::prob_state} that the target model's robustness increases when the last few layers are fine-tuned. It implies that there may exist a potential optimal configuration of $k$ for transfer learning, and we leave the corresponding searching strategies for our future work.

\begin{table*}[ht]
    \center
    \caption{ Ablation studies of \scheme in the scenario of CIFAR-100 $\rightarrow$ CIFAR-10. }
    \begin{tabular} {c c c c c c c c}
        \toprule
        \multicolumn{2}{c}{\multirow{1}{*}{Method}}     & \multicolumn{2}{c}{Case-4}        & \multicolumn{2}{c}{Case-6}        & \multicolumn{2}{c}{Case-8}    \\
        \cmidrule(lr){1-2}                                \cmidrule(lr){3-4}                  \cmidrule(lr){5-6}                  \cmidrule(lr){7-8}
        Source             & Transfer                  & Acc.($\%$)    & Rob.($\%$)        & Acc.($\%$)    & Rob.($\%$)        &  Acc.($\%$)   & Rob.($\%$)    \\
        \midrule
        AT                  & TL                        & 83.22         & 25.23             & 86.92         & 25.38             & 90.82         & 18.54         \\
        AT                  & NEFT                      & 83.72         & 26.29             & 86.87         & 27.95             & 90.92         & 29.97         \\
        AT + FDM            & NEFT                      & 85.51         & 28.47             & 87.79         & 30.94             & 91.30         & 29.34         \\
        \bottomrule
    \end{tabular}
    \label{tab::ablation}
\end{table*}

\begin{table*}[t]
    \centering
    \caption{Comparison of \scheme with LwF and the vanilla method in multiple scenarios.}
    \begin{tabular}{c c c c c c c c c}
        \toprule
        \multirow{2}{*}{Source} & \multirow{2}{*}{Target} & \multirow{2}{*}{Arch.}  & \multicolumn{2}{c}{LwF}       & \multicolumn{2}{c}{Vanilla}       & \multicolumn{2}{c}{CARTL}    \\
                                                                                     \cmidrule(lr){4-5}             \cmidrule(lr){6-7}                  \cmidrule(lr){8-9}
                                &                         &                         & Acc.(\%)      & Rob.(\%)      & Acc.(\%)      & Rob.(\%)          & Acc.(\%)      & Rob.(\%)     \\
        \midrule
        CIFAR-100               & SVHN                    & WRN 34-10 ($k$=6)       & 85.90         & 6.67          & 92.83         & 17.64             & \bacc{93.96}  & \brob{22.21} \\
        CIFAR-100               & GTSRB                   & WRN 34-10 ($k$=6)       & 70.34         & 15.85         & 80.40         & 30.25             & \bacc{83.07}  & \brob{47.34} \\
        CIFAR-10                & SVHN                    & WRN 28-4 ($k$=6)        & 94.32         &  4.68         & \bacc{94.86}  & 11.52             & 94.76         & \brob{21.65} \\
        GTSRB                   & SVHN                    & WRN 28-4 ($k$=6)        & 81.80         &  1.08         & 93.91         &  6.08             & \bacc{94.07}  & \brob{15.26} \\
        \bottomrule
    \end{tabular}
    \label{tab::more_scenarios}
    \vspace{-0.2cm}
\end{table*}

\subsection{Selections for Hyper-parameters}
In this subsection, we evaluate how hyper-parameters of \scheme
affect the accuracy and robustness of the target model.

First, we test the effect of increasing the number of retrained layers.
We report both the target model's accuracy and robustness when fine-tuning the layers of the last $4$, $6$, and $8$ blocks of WRN 34-10, which are denoted as Case-$k$, and $k=4, 6, 8$. When we increase the number of retrained blocks, the accuracy generally rises from $\approx 86\%$ to $\approx 91\%$. 
As for the robustness, \scheme exhibits similar trends to the vanilla method, achieving higher robustness at Case-6.

We further investigate how the hyper-parameter $\lambda$ affects both the accuracy and robustness of the target models. We observe that for all cases, a smaller $\lambda$ helps robustness transfer, especially for the Case-8. 
On the other hand, we can see that reducing $\lambda$ cast slight negative impacts to the accuracy of target models. 

Finally, we evaluate the best choices for the hyper-parameter $\beta$. Recall that in Eq. \eqref{eq::neft_penalty}, we divide weights of fine-tuned layers by their largest singular value while multiplying a scalar $\beta$ for further scaling their Lipschitz constants. 
For example, if we let $\beta=0.4$, the Lipschitz constants of all fine-tuned layers are reduced to about $\beta^2 = 0.16$. We can see that reducing Lipschitz constants significantly improves the target models' robustness from $17.62\%$ to $29.34\%$ but brings a negligible negative impact to the accuracy.

        
        


\begin{figure}
    \center
    \includegraphics[width=\linewidth]{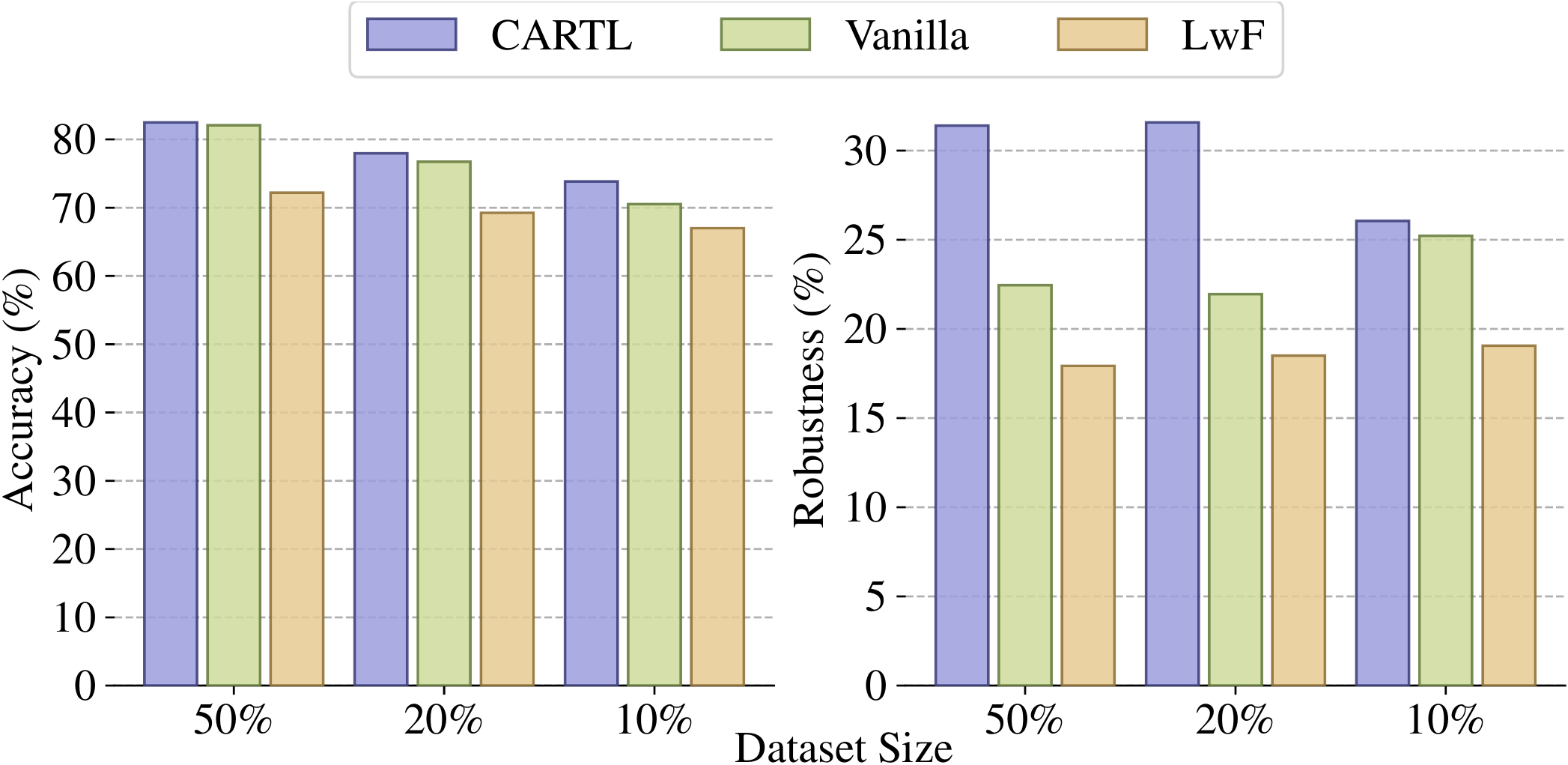}
    \caption{ The accuracy and robustness of target models fine-tuned on CIFAR-10 of different sizes. 
    For comparison, we also train target models with LwF and the vanilla method.
    }
    \label{fig::data_size}
    \vspace{-0.3cm}
\end{figure}

\subsection{Ablation Studies}
To test the effect of each component of \scheme on the accuracy and robustness of the target model, we replace part of the components of \scheme with the vanilla method while training the source models and fine-tuning the target models.
In Table \ref{tab::ablation}, the first row presents the results using the vanilla method, which is used as the baseline. 
Recall that the vanilla method is consisted of simply adversarially training the source model with~\cite{madry_towards_2018}, denoted as AT, and directly fine-tuning the source model on the target domain, denoted as TL.
For the second row, we transfer an adversarially trained model with NEFT, and the last row presents the results achieved by \scheme. 

We can see that fine-tuning the target model with NEFT significantly increases its robustness by observing the first and second rows. 
Besides, FDM further improves the robustness except for the Case-8. We attribute this to the reason that with the increasing number of fine-tuned layers, the effect of constraining the features' distance (\ie, FDM) is reduced. On the other hand, the Lipschitz constraint (\ie, NEFT) plays a more important role when we fine-tune more layers. 
As can be seen from the last column, the robustness increases from $18.54\%$ to $29.97\%$.
Finally, we find that by using FDM, the target model's accuracy slightly rises in all cases.

\subsection{Transfer to Smaller Dataset}
In this subsection, we do additional experiments to test the effect of data size on the target model's performance. For the target domain, CIFAR-10, we fine-tune target models on subsets with the sizes of $50\%$, $20\%$, and $10\%$. To mitigate bias, the number of training data for each label is equal. We summarize and plot the results in Figure \ref{fig::data_size}. Fine-tuning the last six blocks with \scheme maintains better accuracy in all cases, and \scheme provides outstanding robustness except on the extremely small training set. However, \scheme still slightly outperforms the vanilla method in that case.
In comparison, LwF relatively results in the lowest accuracy and robustness in all cases.
Generally speaking, \scheme provides a better accuracy-robustness trade-off for small datasets.






\subsection{Studies on Extra Scenarios}
To demonstrate the generality of \scheme, we conduct experiments in scenarios including transferring from CIFAR-100 to SVHN, from CIFAR-100 to GTSRB, from CIFAR-10 to SVHN, and from GTSRB to SVHN.
According to Table~\ref{tab::more_scenarios}, we can see that the target model fine-tuned with \scheme inherits superior robustness from the source model. Besides, \scheme provides comparable accuracy against the vanilla method. Table~\ref{tab::more_scenarios} demonstrates that \scheme can universally improve the accuracy-robustness trade-off.
As for LwF, it leads to lower accuracy on the natural inputs and fails to guarantee the robustness transfer. In all scenarios, LwF obtains the lowest robustness.




\section{Conclusion \& Future Work}
In this work, we have revealed the trade-off between the robustness and the accuracy of the target model transferred from an adversarially-trained source model. From our observation, we have proposed \scheme, which consists of feature distance minimization and non-expansive fine-tuning, to help the target model inherit more robustness from the source model while maintaining high accuracy. 
We have also found that freezing all batch normalization layers' affine parameters can further improve the transferred robustness.
We hope our work brings insights to enable the following researchers to build a more robust and accurate model in the transfer learning scenario. 

As our future work, we would like to solve the limitation that FDM requires pre-training the source model with an explicitly defined $k$. Other interesting directions include improving the effect of robustness transfer further and considering more security threats against DNNs.

\section*{Acknowledgements}
This work was partially supported by the National Key R\&D Program of China (2020AAA0107701), the NSFC under Grants U20B2049, 61822207, 61822309, 61773310 and U1736205, RGC HK under GRF projects CityU 11217819 and 11217620, RIF project R6021-20, and BNRist under Grant BNR2020RC01013.

\balance

\bibliography{cartl,other}
\bibliographystyle{icml2021}

\clearpage
\appendix

\section{Spectrum Normalization} \label{app::sec::spectrum_norm}
In this section, we briefly recap spectrum normalization \cite{miyato_spectral_2018} for NEFT.

Without loss of generality, we define a linear function $f(\vect{x}) = \vect{W} \vect{x} + \vect{b} $, where $\vect{W} \in \mathbb{R}^{m \times n}$ and $\vect{b} \in \mathbb{R}^{m \times 1}$. The Lipschitz constant of $f$ is upper bounded by its largest singular value $\sigma(\vect{W})$, \ie, the largest eigenvalue of $\vect{W}^T\vect{W}$. Because directly calculating $\sigma(\vect{W})$ is costly, we can use an iterative method to approximate it. To do so, we define two vectors, $\vect{u} \in \mathbb{R}^{m}$ and $\vect{v} \in \mathbb{R}^{n}$, and iteratively calculate $\sigma(\vect{W})$ by
\begin{equation}
    \begin{aligned}
        \sigma(\vect{W})  &= \vect{u}^T \vect{W} \vect{v}, \\
        \vect{v}_{t+1}    &= \vect{W}^T \vect{u}_t / \parallel \vect{W}^T \vect{u}_t \parallel_2, \\
        \vect{u}_{t+1}    &= \vect{W} \vect{v}_{t+1} / \parallel \vect{W} \vect{v}_{t+1} \parallel_2.
    \end{aligned}
    \label{app::eq::spectrum_norm}
\end{equation}
To constrain the Lipschitz constant of $f$ to be one, we divide $\vect{W}$ by $\sigma(\vect{W})$. 

\section{Details of LwF \protect\cite{shafahi_adversarially_2020}} \label{app::sec::lwf}
This section introduces LwF \cite{shafahi_adversarially_2020}, which is considered as one of our baselines. In LwF, Shafahi \etal \cite{shafahi_adversarially_2020} fine-tune all layers of the source model while reducing the difference between the penultimate layer outputs of the target model and the source model. The loss of LwF can be expressed as
\begin{equation}
    \begin{aligned}
        \mathcal{L}_{\text{LwF}} & := \mathcal{L}_{\text{CE}} (f(\vect{x}; \vect{\theta}), y)  \\
                                 & + \lambda_d \cdot \parallel f^{(L-1)} (\vect{x}; \vect{\theta}) - f^{(L-1)}(\vect{x}; \vect{\theta}_0) \parallel_2,
    \end{aligned}
    \label{app::eq::lwf}
\end{equation}
where $\vect{\theta}_0$ is the original parameters of the source model, and $\vect{\theta}$ is the target models' parameter. The hyper-parameter $\lambda_d$ controls the trade-off between the target domain accuracy and the inherited robustness. In Figure \ref{fig::lwf_cmp_ours}, we follow the settings in \cite{shafahi_adversarially_2020} where $\lambda_d = 0.1$, $0.01$, $0.005$, and $0.001$, and we provide detailed results in Table \ref{app::tab::lwf_value}. For other results of LwF (\ie, Table \ref{tab::more_scenarios}, Figure \ref{fig::data_size}), we let $\lambda_d = 0.1$, where it achieves best transferred robustness.

\begin{table} [h!]
    \center
    \caption{Accuracy and robustness of target models transferred from CIFAR-100 to CIFAR-10 using LwF. }
    \begin{tabular} {c | c c }
        \toprule
        $\lambda_d$     &   Acc.($\%$)  &   Rob.($\%$)  \\
        \midrule
        $0.1$           &   74.87       &   17.59       \\
        $0.01$          &   81.45       &   16.67       \\
        $0.005$         &   84.86       &    8.59       \\
        $0.001$         &   89.87       &    0.22       \\
        \bottomrule
    \end{tabular}
    \label{app::tab::lwf_value}
\end{table}

\section{Experiment Settings} \label{app::sec::settings}
In this section, we provide the experiment settings for our evaluations.

\begin{itemize}
    \item We adopt SGD with a momentum of $0.9$ as the optimizer for training all models. The learning rate is initialized as $0.1$ and decays at epoch $40$, $70$, and $90$ by a rate of $0.2$. We train models with a batch size of 128 and set the training epoch as 100. Similar settings are also applied during transfer learning. 
    \item For adversarial training, we utilize PGD-7 to generate adversarial examples during training source models. The $\ell_\infty$-norm constraint (\ie, $\epsilon$) is $8/255$, and the step-size is $2/255$.
    \item We set the number of iterations for the spectrum normalization to be one to avoid introducing extra computational overhead, and the experimental results demonstrate a perfect approximation.
    \item In the model evaluation, we report both the accuracy and robustness of the trained model on the entire test set. Specifically, for the robustness evaluation, we adopt the PGD-100 attack implemented by Foolbox\footnote{https://github.com/bethgelab/foolbox}, an adversarial attack framework, with the step-size of $2/255$ and $\epsilon=8/255$.
    \item When we transfer robust source models to the target domain with \scheme and the vanilla method, we freeze all BN layers' affine parameters, including the feature extractor and the sub-model.
    \item The network architectures used in Section \ref{sec::bn} are detailed in Table \ref{app::tab::bn_arch}.
\end{itemize}

\begin{table}[ht!]
    \center
    \caption{Network architecture configuration of experiments for investigating BN layers' effect. }
    \begin{tabular} {c c c }
        \toprule
        Source      &   Target      &   Arch.       \\
        \midrule
        CIFAR-100   & CIFAR-10      &   WRN 34-10   \\
        CIFAR-10    & GTSRB         &   WRN 28-4    \\
        CIFAR-10    & SVHN          &   WRN 28-4    \\
        \bottomrule
    \end{tabular}
    \label{app::tab::bn_arch}
\end{table}

\end{document}